\documentclass[letterpaper, 10 pt, conference]{ieeeconf}
\IEEEoverridecommandlockouts
\usepackage{cite}
\usepackage{amsmath,amssymb,amsfonts}
\usepackage{algorithmic}
\usepackage{algorithm}
\usepackage{graphicx}
\usepackage{tabularx}
\usepackage{textcomp}
\usepackage{xcolor}
\usepackage{stfloats}
\usepackage{url}
\usepackage{verbatim}
\usepackage{nccmath}
\usepackage{mathtools}
\newcommand{\bb}[1]{\boldsymbol{#1}}

\DeclareMathOperator*{\argmin}{arg\,min}
\begin{document}

\title{Cost Function Estimation Using Inverse Reinforcement Learning with Minimal Observations
}
\author{Sarmad Mehrdad$^{1}$, Avadesh Meduri$^{1}$ and Ludovic Righetti$^{1,2}$
\thanks{This work was in part supported by the National Science Foundation grants 2026479, 2222815 and 2315396 and Wandercraft.}
\thanks{$^1$Tandon School of Engineering, New York University, USA}
\thanks{$^2$Artificial and Natural Intelligence Toulouse Institute (ANITI), France}
}

\maketitle

\begin{abstract}
We present an iterative inverse reinforcement learning algorithm to infer optimal cost functions in continuous spaces. 
Based on a popular maximum entropy criteria, our approach iteratively finds a weight improvement step and proposes a method to find an appropriate step size that ensures learned cost function features remain similar to the demonstrated trajectory features.
In contrast to similar approaches, our algorithm can individually tune the effectiveness of each observation for the partition function and does not need a large sample set, enabling faster learning. We generate sample trajectories by solving an optimal control problem instead of random sampling, leading to more informative trajectories. The performance of our method is compared to two state of the art algorithms to demonstrate its benefits in several simulated environments.
\end{abstract}

\section{Introduction}
Inverse Reinforcement Learning (IRL) and Inverse Optimal Control (IOC) both aim to extract the cost function of a task given the observation(s) of an "expert" or "optimal" task execution \cite{apprentice}. The problem of extracting a cost function from expert observations has several benefits. Primarily, it goes beyond explaining the expert behavior to one environment in a particular task, and tries to explains \textit{why} the task was executed in the observed manner. Moreover, this allows for a generalized behavior of the system facing various environments and inevitable external inputs where simply \textit{mimicking} the expert cannot be fruitful. Because of this, IRL has become attractive for researchers in various fields such as optimal control \cite{IOC-optimalcontrol}, human intent inference \cite{IOC-human-inference}, human movement analysis \cite{panchea2018human} and humanoids \cite{IOC-humanoid} (cf.  \cite{review1, review2} for extended reviews of the topic).

However, extracting an optimized cost function given an expert demonstration is challenging at best. The most important challenge is that there exists multiple cost functions that can explain the observed expert movement, making IRL an ill-posed problem full of local optima. This is in contrast to reinforcement learning (RL) or optimal control (OC), where the cost and task space information is readily available and therefore understood easily. It is also fairly complicated for IRL to sample non-optimal demonstrations for training, and providing expert examples for IRL is not always feasible.

Prior work have tried to solve the IRL problem by either using a maximum margin formulation \cite{maxmargin}, or considering a task related probability distribution to quantify the occurrence of the expert demonstration with highest probability given the cost function \cite{maxent} using maximum entropy principle \cite{jaynes}. Initially, in \cite{maxent}, for small discrete domains dynamic programming was used to estimate the distribution accurately. Further in \cite{relent}, this approach was modified to optimize the embedded maximum likelihood estimation given the relative entropy of the samples. Since then, efforts have been made to generalize these method to higher dimensions and continuous domains using various techniques for estimating the distribution such as approximation \cite{levine} and various sampling methods \cite{mrinal1, bretl, gcl}. 

Although these approaches have shown great potential for learning the cost function, they still suffer from various shortcomings. For example, local sampling around expert demonstration, as used in \cite{mrinal1, mrinal2, bretl}, might fail to contain informative and diverse non-optimal demonstrations to represent task space as a whole. Furthermore, requiring initial sample set can be challenging as well. It either necessitates to use specialized sampling techniques, or to use simpler uniform sampling while risking inaccuracy. Aside from the need for initial samples, most approaches require to use all the samples iteratively for accurate estimation of the trajectory space, which can be computationally demanding. In addition, a majority of IRL methods tend to control  convergence based on probabilistic features such as maximum or relative entropy. Hence, the cost function will not be estimated by how optimal it can generate a new trajectory close to the expert demonstration. 
Some of these issues can be mitigated with IOC approaches, where gradient-based methods are used to solve the problem, hence foregoing the need of trajectory sampling. However, IOC approaches tend to be computationally expensive as they require to solve a bi-level optimization problem which includes an OC problem in its inner loop \cite{IOC-humanoid}.

To address these challenges, this paper proposes an iterative sampling-based IRL algorithm based on maximum entropy principles for continuous task spaces. The approach seeks to iteratively find improvements in the cost estimate while ensuring it would generate trajectories close to the demonstration. We reformulate the probability distribution of the demonstrations by individually tuning the effectiveness of each sample. Moreover, sampling is conducted using an OC solver, where at each iteration we generate new trajectories based on the current cost function estimation. This generates meaningful samples consistent with the the intended use of the cost function and relaxes the need for an ad-hoc sampling strategy.  We demonstrate empirically that our approach leads to significantly faster convergence and higher quality estimates when compared to state of the art methods. 

The remainder of this paper is structured as follows. In Section \ref{preliminaries} we present a brief preliminary to our work, followed by full description of our method in Section \ref{methods}. We present the results for our method in Section \ref{results}, and discussions in Section \ref{disscussions}.

\section{Preliminaries}
\label{preliminaries}

Our method relies on the probabilistic approach of \cite{maxent}, which writes the probability of the expert's demonstration (referred to from now on as the optimal trajectory)
\begin{equation}
    \label{'MaxEnt'}
    P(\tau^\star|\bb{\Bar{\tau}}) = \frac{1}{Z}\exp({-C(\tau^\star)})
\end{equation}
with \textit{partition function}
\begin{equation}
    \label{'partition_function'}
    Z = \exp({-C(\tau^\star)}) + \sum_{\tau_i \in \Bar{\bb{\tau}}}\exp{(-C(\tau_i))}
\end{equation}
where $\tau^\star$ is the optimal trajectory. Observed trajectories are denoted as $\tau = \{\bb{x}_0, \bb{u}_0, \bb{x}_1, \bb{u}_1, \dots , \bb{x}_T\}$ where $\bb{x}_t$ and $\bb{u}_t$ are state and control inputs at time $t$. $T$ is the length of the trajectory. $C(\tau)$ is the cost of  trajectory $\tau$, and $\Bar{\bb{\tau}} = \{\tau_1, \tau_2, \dots, \tau_K\}$ is the set of all observed trajectories, excluding $\tau^\star$. 
With this formulation, trajectories with higher cost are exponentially less likely to occur and vice versa. 

The optimal expert demonstration $\tau^\star$ should then have the highest probability compared to all other possible trajectories. Hence, IRL approaches based on \eqref{'MaxEnt'} aim to maximize the probability of the optimal trajectory, under any set of trajectory observations $\bb{\Bar{\tau}}$

\begin{equation}
    \label{'MaxEntIRL'}
    C^\star = \argmin_{C} -\log(P(\tau^\star|\bb{\Bar{\tau}}))
\end{equation}

A key challenge in such methods is estimating the partition function, i.e. generating a representative set of non-optimal trajectories that will help discriminate cost functions when solving \eqref{'MaxEntIRL'}.
In the following we propose a method for generating trajectories used in the partition function $Z$ that allows the IRL algorithm to have a reasonable understanding of how much each observed trajectory should be incorporated into the partition function. This approach allows for a better representation of the partition function, and even relaxes the need to have all observed trajectories used for the trajectory space representation, hence reducing computational cost.

\section{Methods}
\label{methods}
As mentioned above, we are interested in finding cost functions represented by a linear combination of explicit features. This representation is more constrained than IRL approaches based on neural networks but has two benefits: it allows a seamless integration of the learned cost with optimal control solvers for model-predictive control (MPC) and provides interpretable cost functions, which is important for human movement understanding or human-robot interaction tasks. Further, such cost functions are capable of generating diverse movements and achieving many tasks \cite{crocoddyl}.
We hence search for costs represented as a weighted sum of features

\begin{equation}
    \label{'cost'}
    C(\tau,\bb{w}) = \bb{w}^T\Phi(\tau)
\end{equation}

\begin{equation}
    \label{'feat'}
    \Phi(\tau) = \sum_{t=0}^{T-1} \phi_s(\bb{x}_t,\bb{u}_t)\Delta t + \phi_T(\bb{x}_T)
\end{equation}

\noindent
where $\Phi(\tau)$ is a vector of time integrated features of $\tau$. $\phi_s$ and $\phi_T$ are vectors of stage and terminal cost features, respectively. For brevity, the features of the $i^{\text{th}}$ trajectory is denoted $\Phi_i$, and $\Phi^\star$ is the optimal trajectory's feature vector. 

\subsection{Problem Statement}
Given the linearity assumption of the cost function, the IRL problem becomes 

\begin{align}
    \label{'problem'}
    &\bb{w^\star} = \argmin_{\bb{w}} -\log(P(\tau^\star|\bb{w},\Bar{\bb{\tau}})) \\
    &\text{s.t.} \hspace{3mm} \bb{w} \geq \bb{0}
\end{align}

This problem aims to find weights that maximize the probability of the optimal trajectory for a set of observed trajectories. 
As a consequence, the recovered weights should then produce the same optimal trajectory when used in an OC solver. However, maximizing the probability of the optimal trajectory does not guarantee this. Indeed,
the partition function cannot be computed for all possible trajectories and will be computed from a finite, often small, set of non optimal trajectories. Therefore, there exist trajectories not included in the set that can have higher probability than the demonstration.
This issue is typical of all IRL approaches based on \eqref{'MaxEnt'}.
In this paper, we mitigate this issue by proposing a way to minimize the difference between the estimated costs and the features of the optimal ($\Phi^\star$) and OC trajectory (denoted henceforth as $\Tilde{\Phi}$). 

We propose to solve the problem iteratively by taking small steps $\Delta w$ to improve the cost candidate and propose a procedure to accept such a step at each iteration to ensure it also improves the trajectory that will be generated by the OC solver.

\subsection{Computing a step direction}
We can explicit the minimizing of \eqref{'problem'} as
\begin{equation}
    \label{'argmin_simp'}
    \bb{w}^\star = \argmin_{\bb{w}} -\log \frac{1}{1 + \sum_{\tau \in \Bar{\bb{\tau}}} e^{-\bb{w}^T(\Phi_i - \Phi^\star)}}
\end{equation}

\noindent
Since we seek to find an improvement step, we write the weights at iteration $n+1$ as $\bb{w}_{n+1} = \bb{w}_n + \Delta\bb{w}_n$ which leads to the following problem

\begin{align}
    \label{'delta_w'}
    \Delta\bb{w} &= \argmin_{\Delta\bb{w}} -\log \frac{1}{1 + \sum_{\tau \in \Bar{\bb{\tau}}} \gamma_i e^{-\Delta\bb{w}_t^T(\Phi_i - \Phi^\star)}} \\
    &\text{s.t.}  \hspace{3mm} \Delta\bb{w}_t > -\bb{w}_t
\end{align}
where
\begin{align}
\label{'gamma'}
    \gamma_i = e^{-\bb{w}_t^T(\Phi_i - \Phi^\star)}
\end{align}
can be understood as a weight on each trajectory to give more importance
to non-optimal trajectories that are more probable. This means that trajectories that are farther from optimality will weigh less on the 'changes' in weights and vice versa.
The benefits of assigning a weight to each trajectory when computing the partition function has already been discussed in \cite{gcl}, where an importance sampling argument was used to specifically compute these weights. In our approach, weights come naturally as we search for an improvement direction $\Delta w$.

\subsection{Step Acceptance Method}
Once a step candidate $\Delta \bb{w}$ has been computed by solving problem \eqref{'delta_w'}, we need to 
determine the actual step size to ensure that the change of weight will help find a cost function that will lead to trajectories closer to the optimal one. Reminiscent to line search methods in optimization, we generate a trajectory with an OC solver using $\bb{w}_{t+1} = \bb{w}_t + \alpha\Delta\bb{w}_t$ where $\alpha = 1$, and check if this trajectory is closer to the optimal trajectory in either feature space or cost using the following two merit functions
\begin{align}
    \label{'m1'}
    & m_1(\bb{w}_{t+1}) = m_1(\bb{w}_{t} + \alpha\Delta\bb{w}) =  \frac{1}{2}(\bb{w}^T\Phi^\star - \bb{w}^T\Tilde{\Phi})^2 \\
    \label{'m2'}
    & m_2(\bb{w}_{t+1}) = m_2(\bb{w}_{t} + \alpha\Delta\bb{w}) = ||\Phi^\star - \Tilde{\Phi}||_2
\end{align}
We will accept a step if it leads to an improvement on either of these merit functions. For $m_1$, as we can compute its derivative with respect to $\Delta w$, we use Wolfe conditions to measure acceptable improvement.
For $m_2$ we simply require that it is lower than its value at the previous iteration. We then add the trajectory to $\Bar{\bb{\tau}}$ and iterate again. Otherwise, we decrease $\alpha$ and try again. 
In our algorithm, if we did not find an acceptable step-size after 10 iterations, the algorithm terminates.

The procedure above also explains our approach to sampling, as we add trajectories computed with an OC solver in our dataset. Initially, the algorithm has bad weight estimation and it results in trajectory samplings that are far from optimal. As iterations proceed, generated trajectories become closer to the optimal one. This way, our trajectory sample set will contain both "good" and "bad" non-optimal trajectories. 

\subsection{Sub-sampling Trajectories} \label{sec_subsamp}
We propose to augment our dataset by adding trajectories of reduced length from the current trajectories. For each full length trajectory in the partition function, we create a sub trajectory starting from $t=d$ and ending at $t=T$. We generate several subset of trajectories by using different values of $d\in D$. Our minimization is then adapted to

\begin{equation}
    \label{'delta_w_samp'}
    \Delta\bb{w} = \argmin_{\Delta\bb{w}} \sum_{d \in D} -\theta_d \log \frac{1}{1 + \sum_{\tau \in \Bar{\bb{\tau}}} \gamma_i e^{-\Delta\bb{w}^T(\Phi_{i,d} - \Phi^\star_d)}}
\end{equation}
where $\Phi_{i,d}$ and $\Phi^\star_d$ denote features truncated from $t=d$ and $D$ is the set of timestamps equidistantly chosen between $0$ and $T$. For example, for $N=3$, in addition to the original set we will have two more subsets, containing trajectories with a one-third and two-thirds of their original lengths. To give more importance to longer trajectories, we scale each subset's likelihood by a coefficient $\theta_d  = (T-d+1)/(T+1)$. This can be useful for lower dimensional feature spaces, and lower length trajectories where the trajectory cost can be less sensitive to weight changes. 

\subsection{Regularization}
We also add L1 and L2 regularization to form an \textit{elastic net} regularization, to lower weights magnitude and favor sparsity. The induced sparsity in $\Delta\bb{w}$ causes the most important features to change weight instead of all at the same time. We thus minimize
\begin{align}
    \label{'delta_w_reg'} 
    \nonumber
    \Delta\bb{w} = \argmin_{\Delta\bb{w}} &\sum_{d \in D} -\theta_d \log \frac{1}{1 + \sum_{\tau \in \Bar{\bb{\tau}}} \gamma_i e^{-\Delta\bb{w}^T\Bar{\Phi}_id}}\\ &+ \lambda|\Delta\bb{w}| + \frac{\beta}{2}||\Delta\bb{w}||_2^2
\end{align}

Regularization is useful when feature space is small, and learning process can be challenging especially when opposing features are present as seen in the experiments below. 

\subsection{Moving Window}
Since our approach restricts the steps taken to improve the weights and further scales the importance of each trajectory in the partition function individually, it is not necessary to keep a large history of trajectories to compute the partition function. Indeed, trajectories
far from the optimal one will have a negligible weight.
Therefore we will only use the last $L$ trajectories at each iteration, discarding older trajectories.

This approach is in contrast to several maximum entropy based IRL algorithms which typically require many trajectories to work well.
This further helps reduce computational cost and empirically we notice that it leads to faster convergence as can be seen in Sections \ref{sec_pm_results} and \ref{sec_kuka_results}.

The full method is summarized in Algorithm \ref{'alg'} below.

\begin{algorithm}
    \caption{MO-IRL}
    \label{'alg'}
    \small
    \begin{algorithmic}[1]
        \renewcommand{\algorithmicrequire}{\textbf{Input:}}
        \renewcommand{\algorithmicensure}{\textbf{Output:}}
        \REQUIRE $\tau^\star$
        \ENSURE $\bb{w}_{irl}$
        \STATE $t \leftarrow 0$, $\bb{w}_t \leftarrow [\bb{0.01}]$ \# \texttt{Non-zero small values}
        \STATE $\tau_t = OC(\bb{w}_t)$, $\Bar{\bb{\tau}} = \{ \tau_0 \}$
        \STATE $M_{1t} \leftarrow \infty$, $M_{2t} \leftarrow \infty$
        \STATE $c_1 = 10^{-4}$, $c_2 = 0.9$ \# \texttt{Wolfe Conditions}
        \WHILE{not converged}
            \STATE Find $\Delta\bb{w}$ using \eqref{'delta_w_reg'} \# \texttt{Step Direction}
            \STATE \# \texttt{Step Acceptance Method}
            \STATE $\alpha \leftarrow 1$
            \WHILE{Step not found for 10 trials}
                \STATE $M_1 \leftarrow m_1(\bb{w}_t + \alpha\Delta\bb{w})$ Using \eqref{'m1'}
                \STATE $M_2 \leftarrow m_2(\bb{w}_t + \alpha\Delta\bb{w})$ Using \eqref{'m2'}
                \IF{$M_1$ satisfies Armijo and Wolfe conditions on $m_1$ \\ or $M_2 < M_{2t}$}
                    \STATE $\bb{w}_{t+1} \leftarrow \bb{w}_t + \alpha \Delta\bb{w}$ \# \texttt{Step Found}
                    \STATE $M_{1t} \leftarrow M_1$, $M_{2t} \leftarrow M_2$
                \ELSE
                    \STATE $\alpha \leftarrow \alpha / 4$ \# \texttt{Step Not Found}
                \ENDIF
            \ENDWHILE
            \STATE $\tau_{t+1} = OC(\bb{w}_{t+1})$, $\Bar{\bb{\tau}} \bigcup \{ \tau_{t+1} \}$
            \STATE $t \leftarrow t + 1$
        \ENDWHILE
        
    \end{algorithmic}
    \normalsize
\end{algorithm}

\section{Results}
\label{results}
In this section, we evaluate our approach (called MO-IRL for Minimum Observation IRL) on two different environments and compare it with state of the art approaches. 
The first environment is a simple goal-reaching and obstacle avoidance task with a point mass model. The second environment is a reaching task with obstacle avoidance with a simulated Kuka IIWA 14 robot. We compare our results against the approaches proposed in \cite{mrinal2} and \cite{bretl}, which will henceforth be referred to as PI\textsuperscript{2}-IRL and IS-IRL (for Iterative Scaling), respectively. These methods sample local trajectories based on the method provided in \cite{stomp}. We used Pinocchio \cite{pin} together with the Croccoddyl framework \cite{crocoddyl} and the MiM\_Solver nonlinear SQP solver \cite{sqp} to solve OC problems. We used MuJoCo \cite{mujoco} for the robot simulation. The demonstrated trajectory in all subsequent examples is generated by setting cost weights and using the SQP solver. We do not use the optimal weights anywhere in the tests. 

\subsection{Point Mass Environment}
\subsubsection{Problem setup}
Our first environment is a 2D point mass that reaches a target reaching and obstacle avoidance. The task is to reach the green square, while avoiding the gray circles. The full trajectories are $1.5s$ long for in the one obstacle case, and $2.5s$ seconds long for the other cases with $\Delta t = 0.05$. We use the following cost features: Goal tracking ('G') is a quadratic cost on both state distance to the goal and non-zero velocity,  state and control regularization (Namely 'XReg' and 'UReg') is the squared summation of the corresponding vectors, obstacle avoidance cost for $N_{obs}$ obstacles ('Obs') is zero if the point mass is farther from an obstacle by an activation margin $l$, and otherwise increases quadratically with the distance to the obstacle's center $O_i$. We use the same set of features (excluding UReg) for the terminal costs.

The initial trajectory set for PI\textsuperscript{2}-IRL and IS-IRL were generated by rolling out 20 noisy control inputs in the parametrization of the optimal trajectory. For MO-IRL, however, there is no need for an initial set, and only an OC solution with small non-zero weights will suffice as the 'set' of non-optimal trajectories. For the moving window of the MO-IRL we used $L=1$, meaning that at each iteration only the previously generated trajectory is used for updating the weights. Lastly, we used $N = 20$ sub-samples for the trajectories, and regularized the maximum likelihood estimation by choosing $\lambda = 10^{-6}, \beta = 10^{-2}$.

\subsubsection{Results} \label{sec_pm_results}
\begin{figure}
    \centering
    \includegraphics[width=\linewidth]{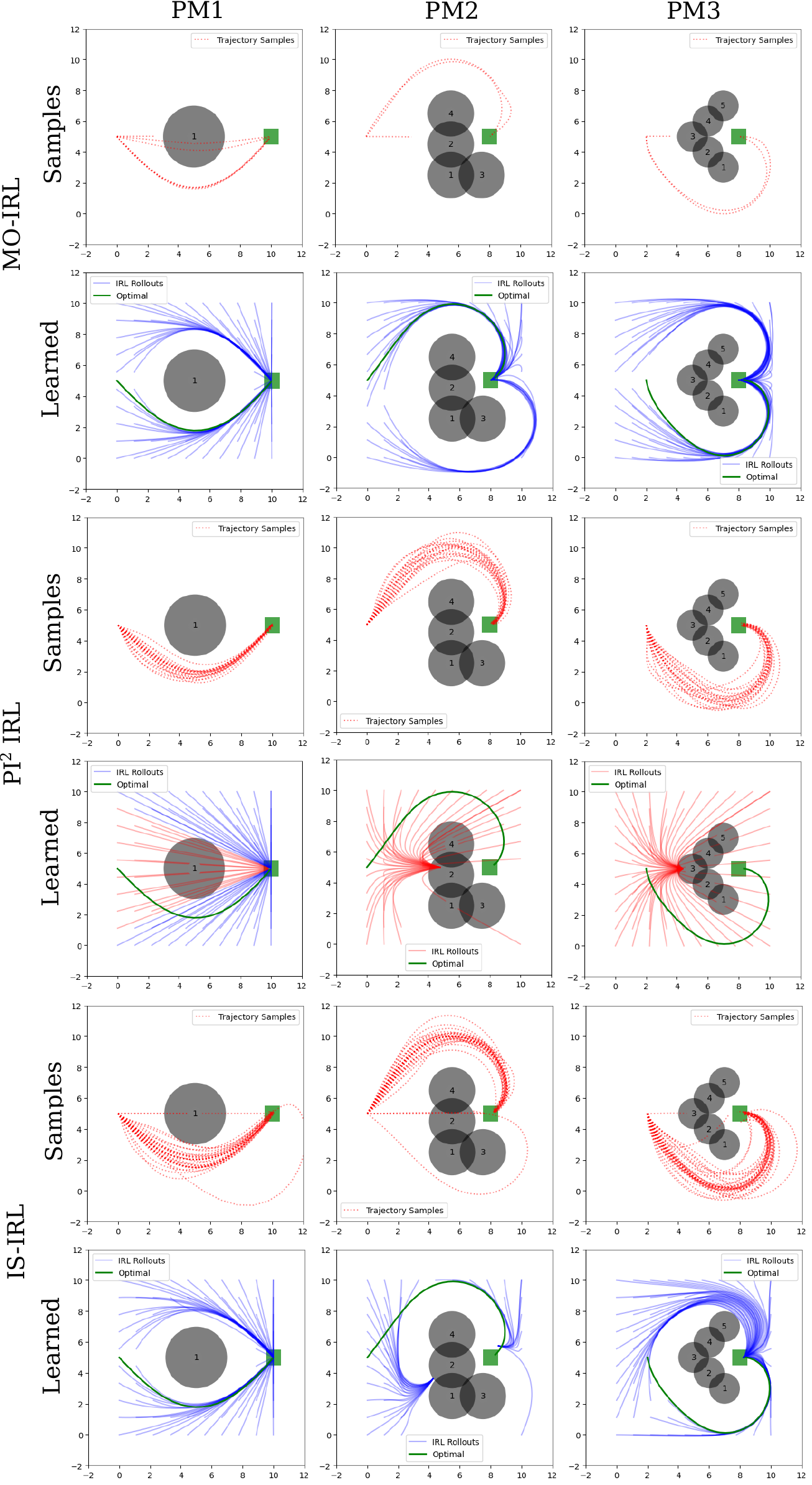}
    \caption{The performance of the IRL algorithms on three point mass environments (one per column). Results are shown in pairs of rows for MO-IRL, PI\textsuperscript{2}-IRL, and IS-IRL. Top row of each pair shows the sampled trajectory set for each algorithm. Bottom rows of pairs show the set of OC rollouts based on the learned cost function. The initial optimal trajectory is shown in green in the bottom row. In the bottom rows, rollouts with obstacle collision are shown in red.}
    \label{fig_arirl_pm}
\end{figure}

Fig.\ref{fig_arirl_pm} shows the performance of MO-IRL in comparison with PI\textsuperscript{2}-IRL and IS-IRL. We test the algorithms in 3 different point mass environments with 1, 4, and 5 obstacles (referred to as 'PM1', 'PM2', and 'PM3', respectively), and show the OC solution for the resulting cost function for the point mass starting from different locations on the map. 
For MO-IRL, convergence happens after 6, 2, and 2 iterations for PM1, PM2, and PM3 respectively, with samplings getting closer to the optimal trajectory faster than the other methods. We can also see, that with changes in the task (starting from other points) the point mass still avoids obstacles and reaches the goal. 

PI\textsuperscript{2}-IRL however, fails to learn a cost that avoids obstacles. This mismatch of goal association comes from the fact that 'XReg' and 'G' features are inherently opposing each other, making it a challenge to tune them well. During optimization, we observed that PI\textsuperscript{2}-IRL fails to lower the 'XReg' weights in the terminal features, which results in the attraction of the point mass towards the starting point when approaching the terminal state. IS-IRL samples trajectories at each iteration by analyzing the entire sampled set, and based on the minimum cost difference of the trajectories, increases the \textit{temperature} of the exponentiated cost in the maximum likelihood estimation process. This tends to render learning more sensitive to smaller changes in the feature space as samples get closer to optimality. Although this approach has the advantage of monitoring the IRL improvement, it has some potential issues. First, the stopping criteria only rests on temperature passing a threshold, and not how the weights converge. As a result, IS-IRL will increase the weights to much higher values than needed by the last iterations, losing sensitivity to some features, which can lead to some issues. For example, we see in Fig.\ref{fig_arirl_pm} that in PM2, all trajectories starting behind the obstacles converge to a local minimum and never reach the goal. Indeed, the loss of sensitivity to certain features led the algorithm to give too large weight on the obstacle avoidance costs to the detriment of target reaching. Secondly, IS-IRL intends to change the effectiveness of the sampled trajectories, but it does so by changing the temperature for the whole set, meaning that the entire set is needed to form the partition function. This can be computationally heavy as iterations proceed. Conversely, in our method the influence of the trajectories is individually tuned given the previous weights, which relaxes the need for having the full set at all times.

\begin{figure}
    \centering
    \includegraphics[width=\linewidth]{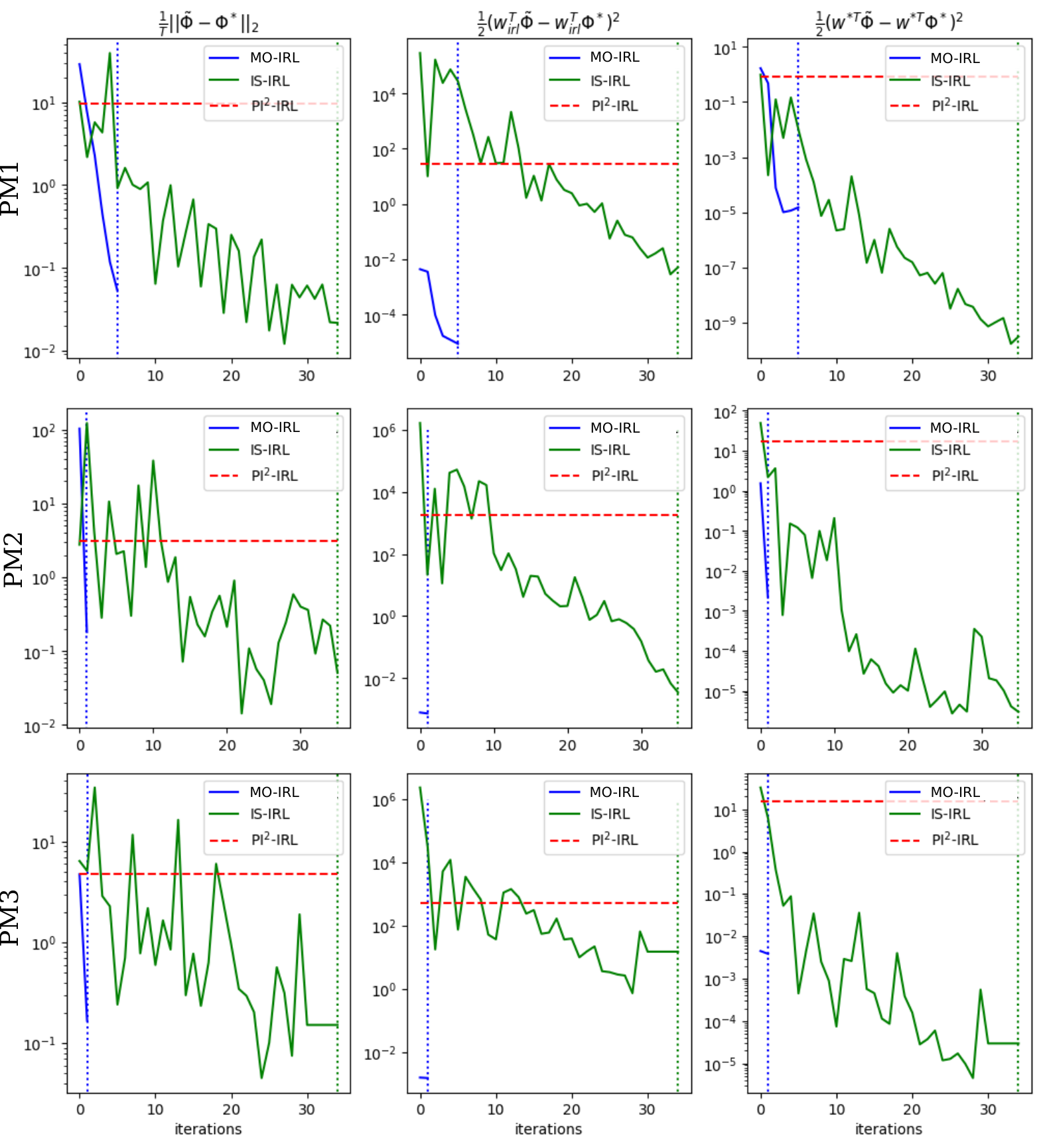}
    \caption{Performance of MO-IRL and IS-IRL throughout their iterations. First column indicates the deviation of the resulting trajectory from optimality, second column shows the cost differences of the optimal and the solved trajectory given the estimated weight set, and the third column shows the cost difference of the optimal and solved trajectory given the optimal weights at each iteration. Vertical dashed lines show the termination of associated algorithms. PI\textsuperscript{2}-IRL's performance is shown by a horizontal red dashed line.}
    \label{fig_pm_iter}
\end{figure}

Fig.\ref{fig_pm_iter} shows the performance of the algorithms on the point mass tests. MO-IRL concludes the training much earlier than IS-IRL, even though it terminates on a higher trajectory costs. IS-IRL usually starts the iterations where PI\textsuperscript{2}-IRL finishes the learning, which is expected since without iterative scaling of the partition function, IS-IRL and PI\textsuperscript{2}-IRL are practically very close in nature. Table \ref{tab_results_PM}, shows the costs of the trajectories generated from the learned weights from each algorithm using the original optimal cost function. Our approach produces a trajectory with a cost closer to the optimal cost. The computation time of the IRL algorithms are shown in Table \ref{tab_duration_PM}, which shows that MO-IRL is fast. We implemented each algorithm in a similar way but our results need to be taken with care as it is always possible to improve an implementation. 

\begin{figure*}
    \centering
    \includegraphics[width=0.8\linewidth]{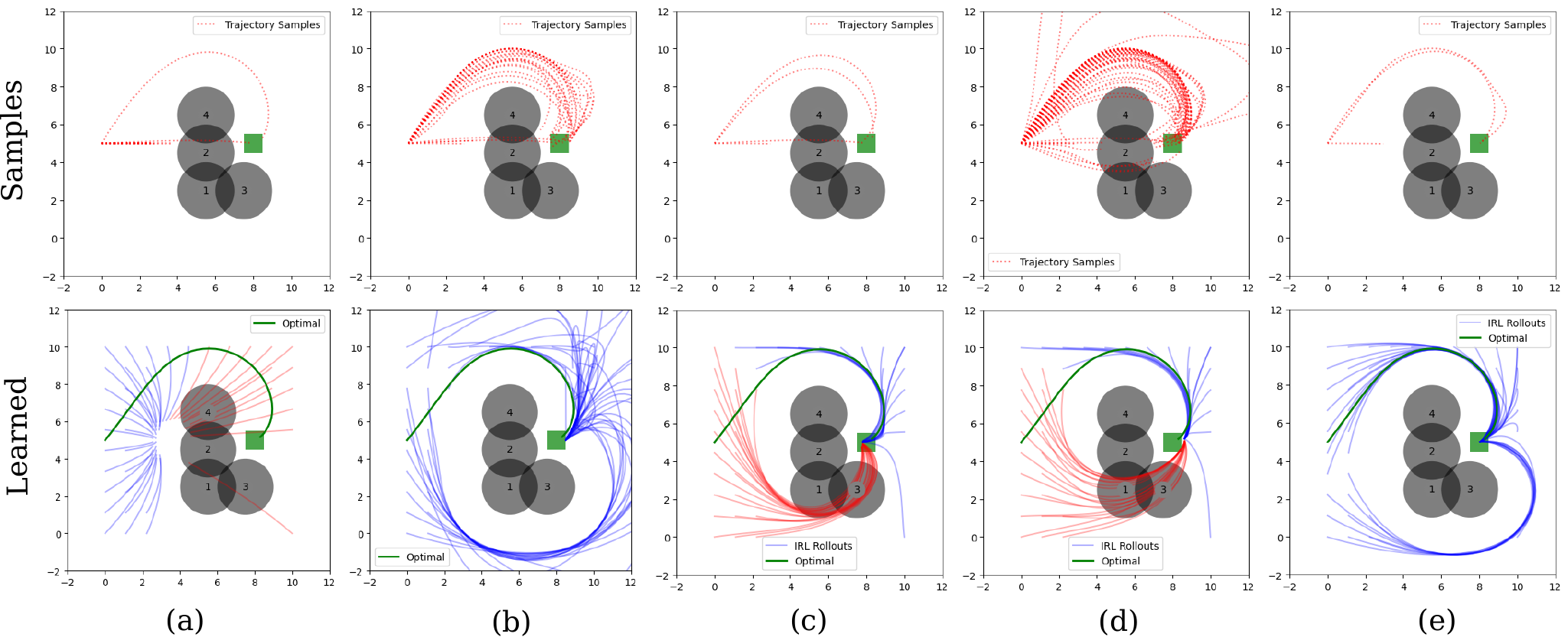}
    \caption{Comparison of each improving step for modifying the IRL performance. \textbf{(a)} shows the algorithm without any step acceptance, regularization, and sub-sampling. \textbf{(b)} is the same as (a) with sub-sampling. \textbf{(c)} shows (b) with added step acceptance method. \textbf{(d)} is (b) but with regularization. \textbf{(e)} shows the full version with sub-sampling, regularization, and step acceptance method. }
    \label{fig_trans_pm}
\end{figure*}

To better understand the importance of each part of the algorithm, we  conducted experiments with PM2 with versions of the algorithm that did not include the step acceptance strategy, the regularization or sub-sampling. The results are summarized in Fig. \ref{fig_trans_pm}. We notice that without either of these approaches, the algorithm fails completely at recovering a good cost. Adding sub-sampling leads to some trajectories reaching the target. Adding the step acceptance strategy further leads to trajectories that always reach the target several are colliding with the obstacles. Regularization similarly improves the behavior of the algorithm, with similar obstacle collision issues. The best results are found when everything is used.

\begin{table}[b]
    \caption{Point Mass costs given optimal cost function}
    \vspace{-5mm}
    \begin{center}
        \begin{tabularx}{\linewidth} { 
          | >{\raggedright\arraybackslash}X 
          | >{\centering\arraybackslash}X 
          | >{\centering\arraybackslash}X 
          | >{\centering\arraybackslash}X
          | >{\centering\arraybackslash}X | }
        \hline
            Cost $\times 10^4$ & \small $\bb{w}^{\star T}\Phi_{MO}$ & $\bb{w}^{\star T}\Phi_{IS}$ & $\bb{w}^{\star T}\Phi_{PI^2}$ & $\bb{w}^{\star T}\Phi^\star$ \normalsize\\ \hline
            PM1 & $ 0.5085$ & $0.5121$ & $1.8025$ & $0.5036$ \\ \hline
            PM2 & $0.6061$ & $6.0114^+$ & $6.3888^+$ & $0.5091$ \\ \hline
            PM3 & $0.5613$ & $0.5779$ & $6.0938^+$ & $0.4970$\\ \hline
            \multicolumn{4}{l}{$^+$Did not converge to target.}
        \end{tabularx}
    \label{tab_results_PM}
    \end{center}
\end{table}


\begin{table}[b]
    \caption{Point Mass IRL computation time (sec)}
    \vspace{-5mm}
    \begin{center}
        \begin{tabularx}{\linewidth} { 
          | >{\raggedright\arraybackslash}X 
          | >{\centering\arraybackslash}X 
          | >{\centering\arraybackslash}X 
          | >{\centering\arraybackslash}X | }
        \hline
            \hspace{5mm} & PM1 & PM2 & PM3 \\ \hline
            MO-IRL & $22.22$ & $10.14$ & $19.37$ \\ \hline
            IS-IRL & $34.53$ & $86.02$ & $92.68$ \\ \hline
            PI\textsuperscript{2}-IRL & $4.16$ & $9.58$ & $5.96$ \\ \hline
            \multicolumn{4}{l}{$^\star$Results are achieved under the same setting on a single device, with} \\
            \multicolumn{4}{l}{an Intel Core i7-10750H CPU @ 2.60GHz}
        \end{tabularx}
    \label{tab_duration_PM}
    \end{center}
\end{table}

\subsection{Robot Simulation}
\subsubsection{problem Setup and Initialization}
In this example, we study a goal reaching task with obstacle avoidance for a simulated torque-controlled 7-DOFs Kuka iiwa robot.
We use the same cost features (albeit in a higher dimensional space) and do the following modifications. The goal cost is computed in end-effector frame. We further create capsules around the last four links of the robot and use the distance between the capsules and the obstacles to compute obstacle feature costs (one feature per pair of capsule / obstacle).
The full trajectories are one second long, with $\Delta t = 0.01$. The environment consists of $N_{obs} = 4$ box obstacles that the robot has to maneuver around in order to get to the green sphere target. For stage cost, there are 19 features (16 for obstacle avoidance, XReg, UReg, and G), and 18 for the terminal (All but UReg), making the total of 37 cost features. The initial trajectory set for PI\textsuperscript{2}-IRL and IS-IRL were generated by 20 noisy DMP rollouts. We noticed that when the feature numbers are relatively high, it is more beneficial to the learning if the weights are bounded as mentioned in \ref{sec_subsamp} so that no \textit{overweighting} would occur. Although this weight binding is a helpful improvement for MO-IRL, it is necessary for IS-IRL since it weighs the obstacle avoidance features high and forgets other cost features, resulting in robot not moving at all. Hence, for the results presented in this section, we used weight bounds for all approaches ($\bb{1} \geq \bb{w}_t \geq \bb{0}$). For MO-IRL, since the cost features are abundantly large in number, it suffices to have $L=1$ and $N=1$, and since the weights are bounded, regularization can also be omitted.

\subsubsection{Results} \label{sec_kuka_results}
\begin{figure}
    \centering
    \includegraphics[width=\linewidth]{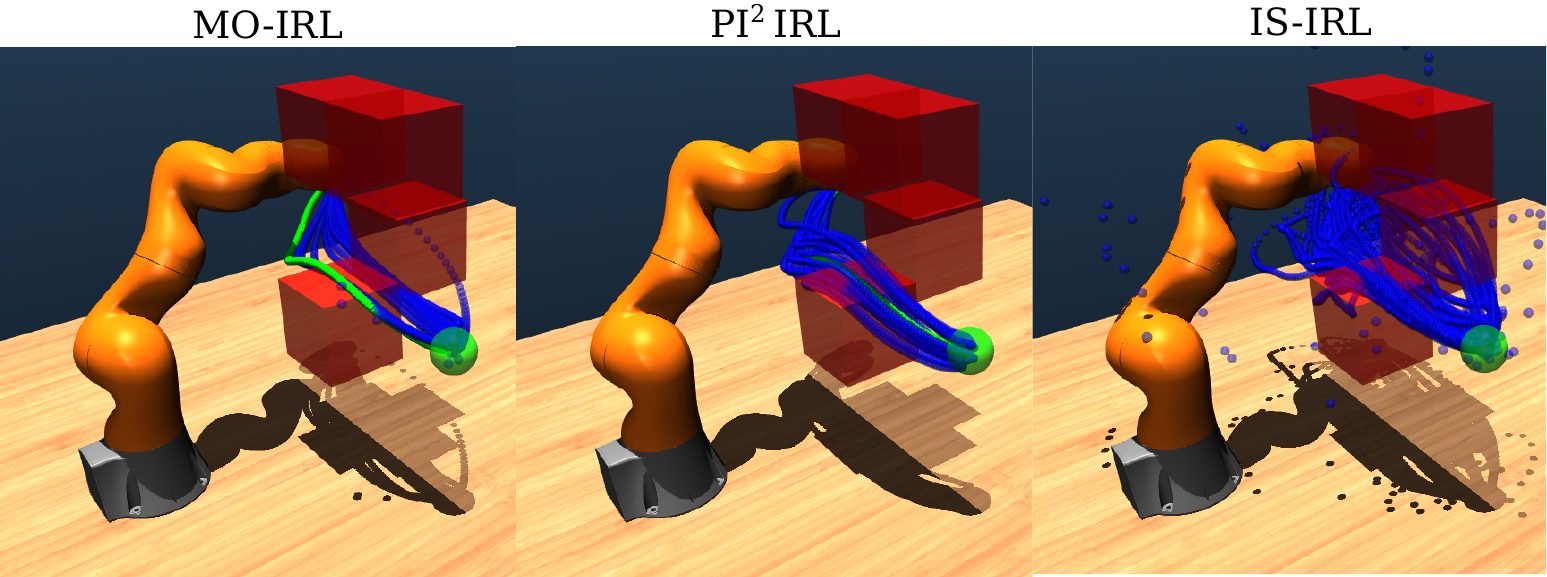}
    \caption{Sampled trajectory sets for the IRL approaches. MO-IRL carries out a different way of sampling that is not necessarily local to the optimal trajectory. PI\textsuperscript{2} IRL uses only the initial noisy local rollouts. IS-IRL concludes the iterations with a trajectory set containing both initial local rollouts, and OC's solutions. The green path seen on the left figure is the optimal trajectory.}
    \label{fig_kuka_sampling}
\end{figure}

Fig.\ref{fig_kuka_sampling} portrays the sampling set for IRL performances in the Kuka simulation. It can be seen that the trajectory sampling of MO-IRL is significantly different from that of the other approaches. The samples needed for MO-IRL are not local unlike the other samplings seen for PI\textsuperscript{2}-IRL and IS-IRL. In addition, due to this particular sampling, MO-IRL can converge with a lower number of trajectory samples. For PI\textsuperscript{2}-IRL the number of samples does not change (20 trajectories). For IS-IRL, this number rose to 55, and for MO-IRL, total sample count was only 14. Although the samples for MO-IRL are not as diverse as the ones for IS-IRL, there are trajectories that have gone through the obstacles that could not have been achieved by local rollouts. This ability of MO-IRL to generate informative bad demonstrations is extremely important for exposing the robot to states that it should learn not to go towards. For example, in the initial stage where there are only $\tau_0$ and $\tau^\star$ sampled, MO-IRL optimize a cost that create trajectories going straight to the goal. This results in sampling a trajectory that hits the obstacles along its way, gaining more knowledge about the environment in subsequent iterations.

\begin{figure}
    \centering
    \includegraphics[width=\linewidth]{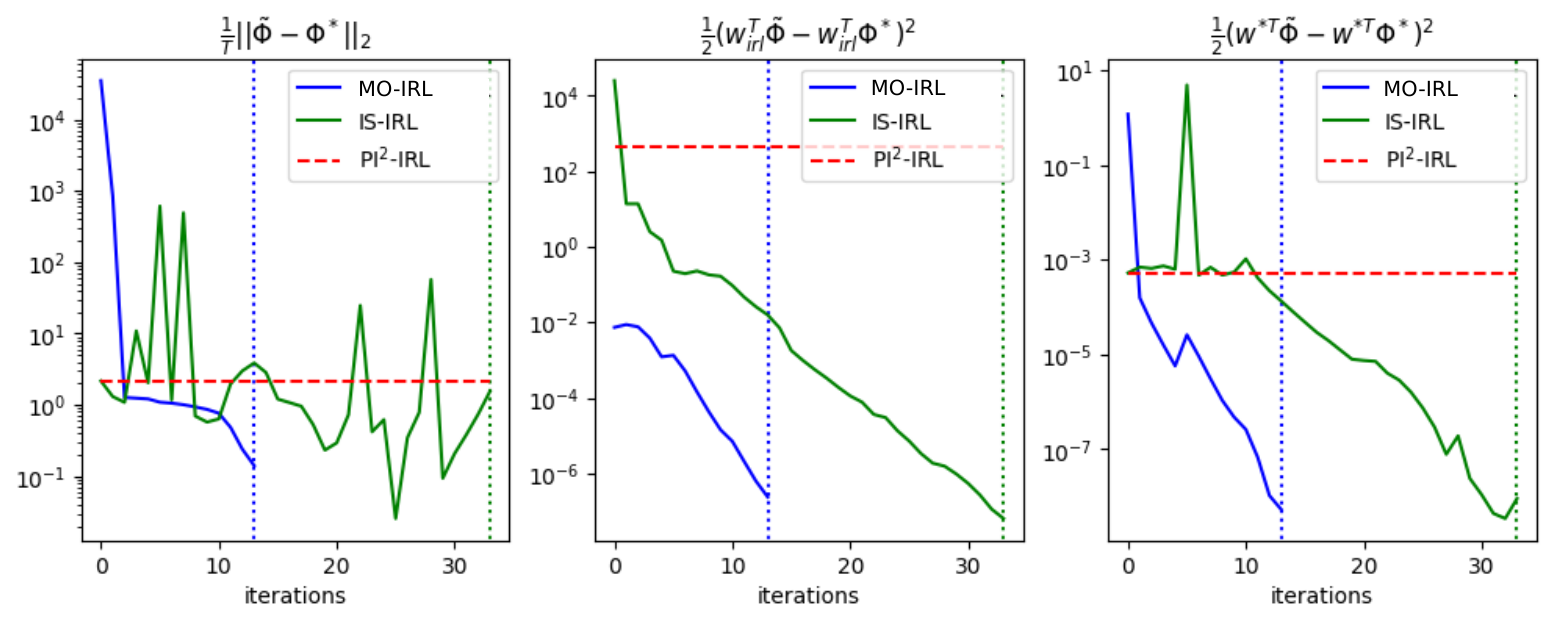}
    \caption{Iterative performance comparison between MO-IRL and IS-IRL. PI\textsuperscript{2}-IRL's performance is indicated by the horizontal dashed red line.}
    \label{fig_kuka_iter}
\end{figure}

We compare the iterative performance of the algorithms for robot simulation in Fig.\ref{fig_kuka_iter}. Similar to point mass tests, MO-IRL terminates after a lower number of iterations with higher optimal trajectory cost difference in comparison to IS-IRL. Unlike in the point mass tests however, MO-IRL produces the most optimal trajectory in terms of the features (left sub-figure). Moreover, since we used only the latest trajectory in MO-IRL, the computation for weight change is less demanding, and the convergence happened twice as fast as IS-IRL.

\begin{figure}
    \centering
    \includegraphics[width=\linewidth]{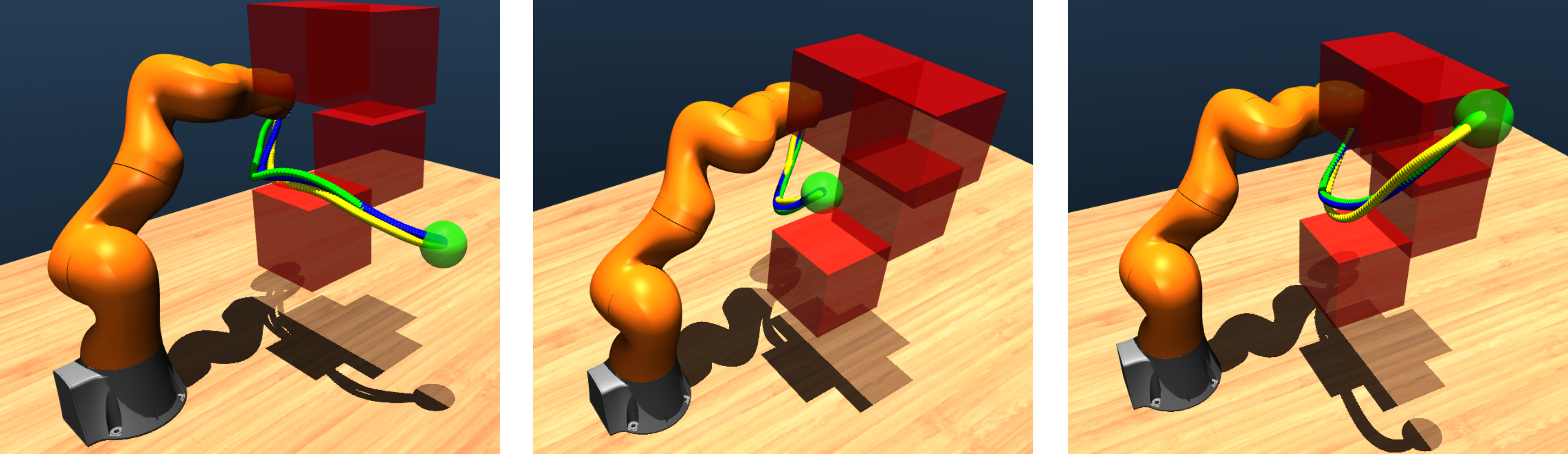}
    \caption{Comparison of robot behavior in the same obstacle environment with different goal locations. MO-IRL, IS-IRL, and optimal solution given the desired weights, are shown in blue, yellow, and green respectively.}
    \label{fig_kuka_same}
\end{figure}

Fig. \ref{fig_kuka_same} shows how the robot conducted the task given the learned weights. PI\textsuperscript{2}-IRL in this test case performed poorly, as the same failure occurred for tuning the XReg against G, therefore robot stands mostly still. For that reason we cannot portray any movement resulting from PI\textsuperscript{2}-IRL in Fig. \ref{fig_kuka_same}. MO-IRL and IS-IRL however showed good performance for the same task and when the goal position was changed. Apart from the environment with the same goal position (left sub-figure) where IS-IRL slightly grazes the lower obstacle, both algorithms showed nearly the same behavior. 

\begin{figure}
    \centering
    \includegraphics[width=0.8\linewidth]{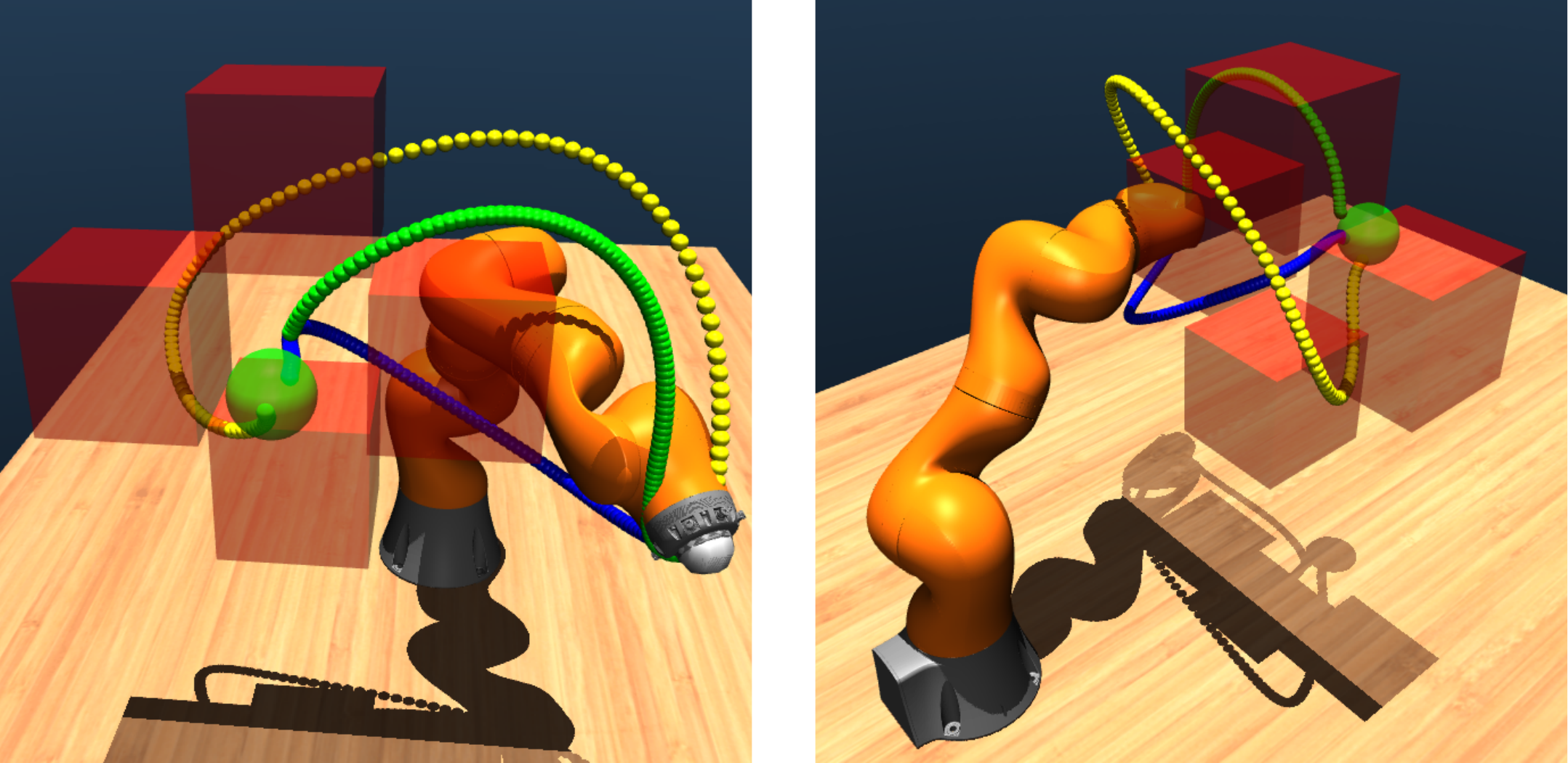}
    \caption{Comparison of robot behavior in different challenging obstacle environment and goal location. The images are for the same test from different angles for better visibility. The color-coding is the same as in Fig. \ref{fig_kuka_same}.}
    \label{fig_kuka_diff}
\end{figure}

To empirically investigate how the learned cost would fare in a different environment, we moved the obstacles and the goal in such a way that the robot needs to needle through the obstacles to get to the target. Fig. \ref{fig_kuka_diff} shows the results for this experiment. Like in the previous environment, the robot shows performance with lower cost closer to the optimal one. However, we notice that the resulting behavior is different from the one generated with the optimal cost. The optimal cost (which we tuned for the first environment) generates a trajectory that hits the obstacles while the trajectory generated by our approach is different and does not hit the obstacles. We hypothesize that since our approach is designed to find small weight changes $\Delta w$, it tends to find smaller, well-balanced weights overall which might lead to better balanced feature selection. This will need, however, to be further investigated.

\begin{table}[b]
    \caption{Robot Simulation costs given optimal cost function}
    \vspace{-5mm}
    \begin{center}
        \begin{tabularx}{\linewidth} { 
          | >{\raggedright\arraybackslash}p{0.8in}
          | >{\centering\arraybackslash}X 
          | >{\centering\arraybackslash}X 
          | >{\centering\arraybackslash}X 
          | >{\centering\arraybackslash}X | }
        \hline
        
            Cost $\times 10^4$ & \small $\bb{w}^{\star T}\Phi_{MO}$ & $\bb{w}^{\star T}\Phi_{IS}$ & $\bb{w}^{\star T}\Phi_{PI^2}$ & $\bb{w}^{\star T}\Phi^\star$ \normalsize\\ \hline
            Same Task & $ 8.635$ & $ 8.949$ & $322.703^+$ & $7.607$ \\ \hline
            Changed Env. & $365.221$  & $394.249$  & $380.1662^+$  & $182.584$ \\ \hline
            Duration & $22.39s$ & $50.61s$ & $100.41s$ & N/A \\ \hline
            \multicolumn{4}{l}{$^+$Did not converge to target.}
        \end{tabularx}
    \label{tab_results_kuka}
    \end{center}
\end{table}

\section{Discussions} 
\label{disscussions}
We presented the MO-IRL algorithm, which iteratively improves the estimate of the optimal cost function by representing the trajectory space with minimal samples. We have shown that the approach can outperform state of the art IRL algorithms. Our method assigns individual tuning to trajectory samples for the partition function at each iteration depending on the most recent weight estimation with respect to the optimal trajectory. This means that any subset of samples in the set using this method can provide a reasonable space representation, relaxing the need for subjective tunings in the partition function, or necessity of including all samples. Due to the independence of MO-IRL to the whole sampled trajectory set, even one trajectory can be sufficient for learning the weights properly. This helps both convergence and computational complexity. Moreover, MO-IRL does not rely on local samplings and consequently does not require a policy parametrization. Instead, the samples are drawn from the OC solver at each iteration where a new cost function is estimated. 

In the future, we aim to exploit the algorithm and its advantageous computational complexity towards online cost learning and MPC implementations. This would be beneficial for human-in-the-loop scenarios where an expert is presently trying to interactively teach a task to a robot. 

\bibliographystyle{IEEEtran}
\bibliography{ref}
\end{document}